%% file: main.tex
\documentclass[conference]{IEEEtran}
\pdfoutput=1
\usepackage{times}
\usepackage[numbers]{natbib}
\usepackage{multicol}
\usepackage{multirow}
\usepackage{colortbl}
\usepackage{booktabs}
\usepackage[bookmarks=true,hidelinks]{hyperref}
\let\labelindent\relax
\usepackage{enumitem}
\usepackage[dvipsnames,svgnames]{xcolor}  
\usepackage[utf8]{inputenc}
\usepackage{graphicx}
\usepackage{epsfig}
\usepackage{bibentry}
\usepackage{textcomp}
\usepackage{hhline}
\usepackage{epstopdf}
\usepackage{mathptmx}
\usepackage{times}
\usepackage{amsmath}
\usepackage{amssymb}
\usepackage{stfloats}
\usepackage{tikz}
\usepackage{pgfplots}
\usepackage{multicol}
\usepackage{multirow}
\usepackage{wrapfig}
\usepackage{caption}
\usepackage{subcaption}
\usepackage{float}

\usepackage{braket}
\usetikzlibrary{automata,positioning,arrows,arrows.meta,fit,decorations.pathreplacing}
\usepackage{lipsum}  
\usepackage{siunitx}

\usepackage{adjustbox}
\usepackage{array}
\usepackage{siunitx}

\usepackage{ifthen}
\usepackage{xstring}
\usepackage{calc}
\usepackage{pgfopts}
\usepackage{tikz-uml}
\usepackage{tikzit}
\input{sample.tikzstyles}

\newcolumntype{R}[2]{%
	>{\adjustbox{angle=#1,lap=\width-(#2)}\bgroup}%
	l%
	<{\egroup}%
}

\DeclareMathAlphabet{\mathcal}{OMS}{cmsy}{m}{n}

\usepackage{blindtext}
\usepackage{varwidth}
\usepackage{algorithm,algpseudocode}

\pgfplotsset{compat=newest}
\usepgfplotslibrary{units}
\usetikzlibrary{arrows}
\usetikzlibrary{angles, quotes}
\usepackage{siunitx}
\newlength{\figwidth}          
\newlength{\figheight}         
\tikzset{
	block/.style={
		draw, 
		rectangle, 
		minimum height=1.5cm, 
		minimum width=3cm, align=center
	}, 
	line/.style={->,>=latex'}
	every edge/.style={ 
		draw,
		->,>=stealth’, 
		auto,
		semithick}}

\newcommand{\eq}[1]{Eq.~\eqref{#1}}
\newcommand{\fig}[1]{Fig.~\ref{#1}} 

\newcommand{\sect}[1]{Section~\ref{#1}}
\newcommand{\alg}[1]{Algorithm~\ref{#1}}

\DeclareMathOperator{\EX}{\mathbb{E}}
\newcommand{\sr}[1]{\mathbf{s}^\textrm{r}_{#1}}
\newcommand{\sv}[1]{\mathbf{s}^\textrm{v}_{#1}}
\newcommand{\svi}[3]{\mathbf{s}^{\textrm{v}_{#2}^{#3}}_{#1}}

\newcommand{\act}[1]{\mathbf{a}_{#1}}
\newcommand{\rew}[1]{r_{#1}}

\newcommand{\Doff}[1]{\mathcal{D}^\textrm{off}}
\newcommand{\D}[1]{\mathcal{D}_{#1}}
\newcommand{\Drec}[0]{\mathcal{D}_\textrm{rec}}
\newcommand{\Nr}[0]{N_\textrm{rec}}

\newcommand{\s}[1]{\mathbf{s}_{#1}}
\newcommand{\tr}[1]{e_{#1}}
\newcommand{\tri}[2]{e_{#1}^{#2}}
\newcommand{\tauhysr}[1]{\tau^\textrm{HySR}_{#1}}
\newcommand{\tauhis}[1]{\tau^\textrm{HiS}_{#1}}

\newcommand{\taurec}[1]{\tau^{\textrm{v}^\textrm{rec}_{#1}}}

\definecolor{linkcolor}{HTML}{648EB0}
\newcommand{\link}[2]{\href{#1}{\textcolor{linkcolor}{#2}}}

\begin{document}

\title{Hindsight States:\\Blending Sim \& Real Task Elements for\\Efficient Reinforcement Learning}

\author{Simon Guist, Jan Schneider, Alexander Dittrich, Vincent Berenz, Bernhard Schölkopf and Dieter Büchler%
\thanks{This work was supported by the Max Planck Institute for Intelligent Systems.}
\thanks{All authors are affiliated with the MPI for Intelligent Systems, Max-Planck-Ring 4, 72076 Tübingen, Germany.}%
}

\IEEEoverridecommandlockouts

\maketitle

\input{his_images}
\begin{abstract}

\input{0_abstract}

\end{abstract}

\IEEEpeerreviewmaketitle

\section{Introduction}
\label{sec:intro}
\input{1_intro}

\section{Related Work}
\label{sec:related_work}
\input{2_related_work}

\section{Method}
\label{sec:method}
\input{3_method}

\section{Experiments}
\label{sec:experiments}
\input{4_experiments_new}

\section{Conclusion and Future Work} 
\label{sec:conclusion}
\input{5_conclusion}

\bibliographystyle{plainnat}
\bibliography{refs}

\clearpage

\newpage

\section{Appendix} 
\label{sec:appendix}
\input{99_appendix}

\end{document}

%% file: sample.tikzstyles
\tikzstyle{circle standard}=[fill=white, draw=black, shape=circle, tikzit fill=white, tikzit draw=black, tikzit shape=circle, minimum size=0.8cm, inner sep=0pt]
\tikzstyle{red circle}=[fill=white, draw={rgb,255: red,191; green,0; blue,64}, shape=circle, thick, minimum size=0.8cm, tikzit draw={rgb,255: red,191; green,0; blue,64}, tikzit fill=white, tikzit shape=circle, inner sep=0pt]
\tikzstyle{blue circle}=[fill=white, draw={rgb,255: red,22; green,8; blue,180}, shape=circle, tikzit shape=circle, tikzit draw={rgb,255: red,22; green,8; blue,180}, tikzit fill=white, minimum size=0.8cm, inner sep=0pt, thick]

\tikzstyle{arrow}=[->, fill=none, tikzit fill=none]
\tikzstyle{dashed arrow}=[->, dashed]
\tikzstyle{blue arrow}=[->, draw={rgb,255: red,22; green,8; blue,180}, tikzit draw={rgb,255: red,22; green,8; blue,180}, fill=none, tikzit fill=white, thick]
\tikzstyle{blue dashed arrow}=[->, dashed, draw={rgb,255: red,22; green,8; blue,180}, thick]
\tikzstyle{dashed line}=[-, dashed]

%% file: his_images.tex
\begin{figure*}
	\centering

    \subfloat[Pushing]{
    \centering\scriptsize%
    \hspace{-.25cm}
    \includegraphics[width=0.23625\textwidth]{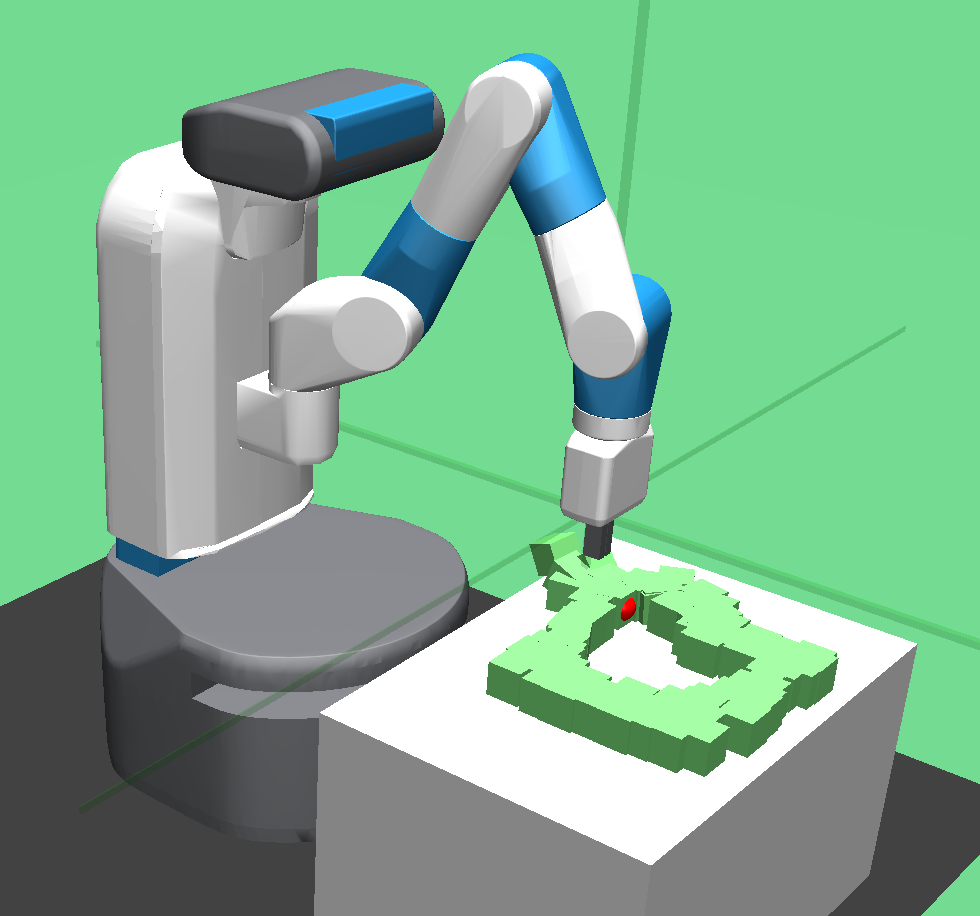}
    \label{sfig:image_push}
}
\subfloat[Sliding]{
    \centering\scriptsize%
    \hspace{-.25cm}
    \includegraphics[width=0.2512125\textwidth]{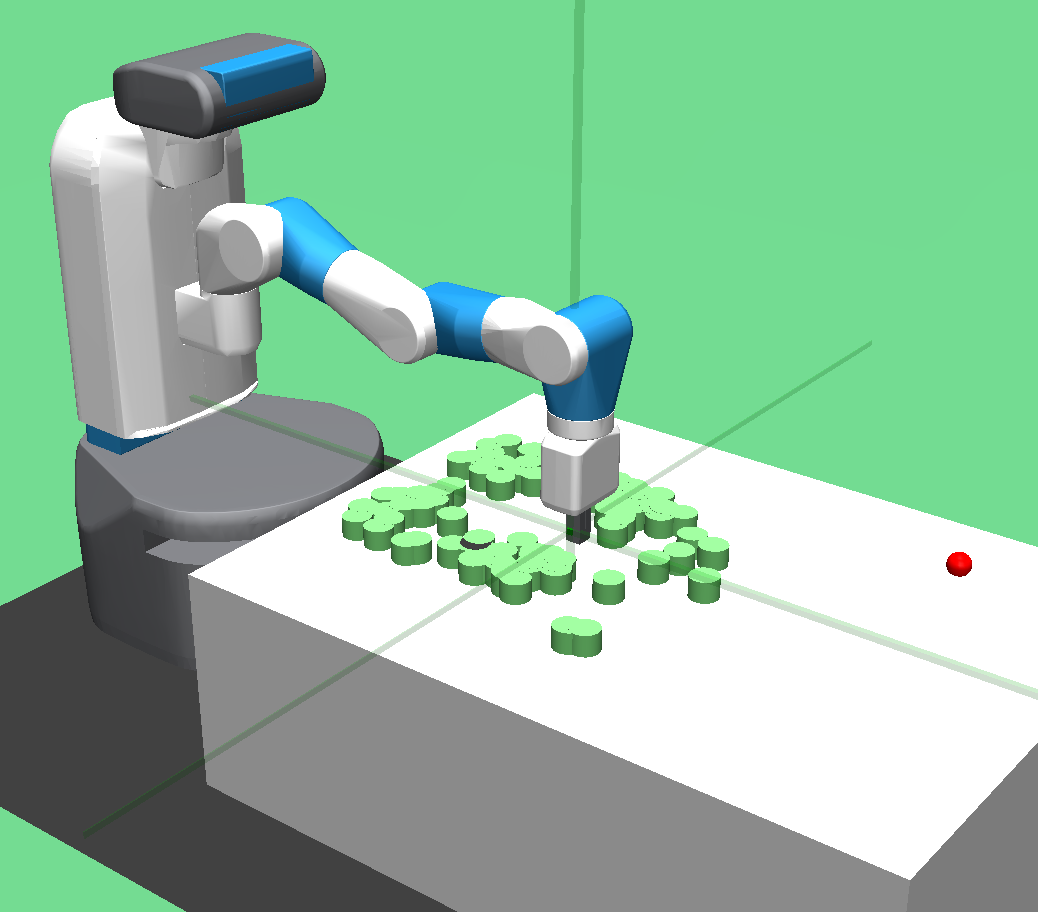}
    \label{sfig:image_slide}
}
\subfloat[Simulated robot table tennis]{
    \centering\scriptsize%
    \hspace{-.25cm}
    \includegraphics[width=0.2905\textwidth]{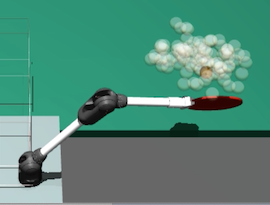}
    \label{sfig:image_table_tennis_sim}
}
\subfloat[Real robot table tennis]{
    \centering\scriptsize%
    \hspace{-.25cm}
    \includegraphics[width=0.152\textwidth]{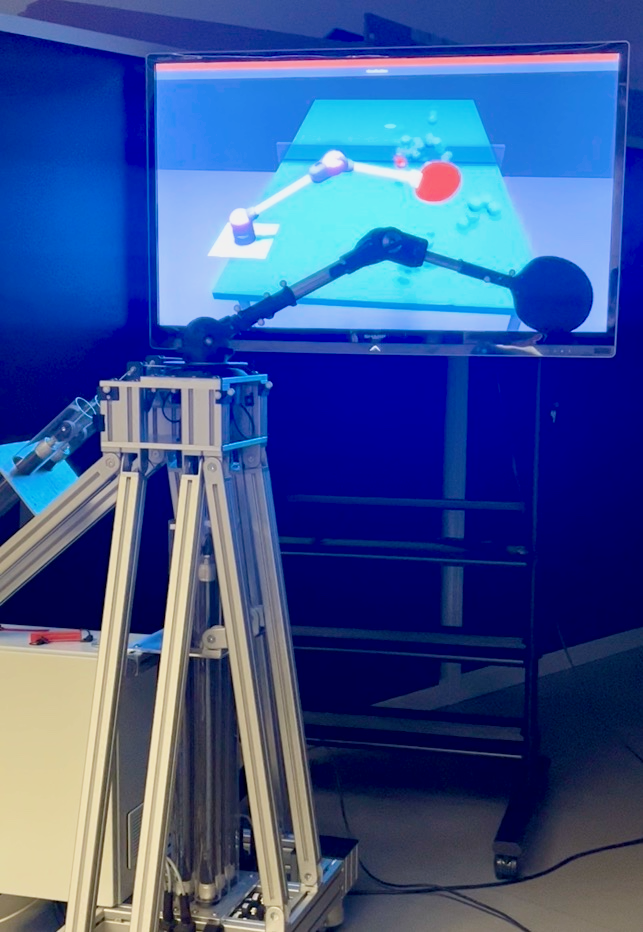}
    \label{sfig:image_table_tennis_real}
}

	\caption{
		Visualization of HiS applied to the tasks considered in this work. 
		Rather than training with a single object, HiS uses multiple virtual objects in parallel to generate more data and experience higher rewards with increased probability early on in the training.
	}
	\vspace{-10pt}
	\label{fig:fetch_tasks}
\end{figure*}

%% file: 0_abstract.tex
Reinforcement learning has shown great potential in solving complex tasks when large amounts of data can be generated with little effort.
In robotics, one approach to generate training data builds on simulations based on dynamics models derived from first principles.
However, for tasks that, for instance, involve complex soft robots, devising such models is substantially more challenging.
Being able to train effectively in increasingly complicated scenarios with reinforcement learning enables to take advantage of complex systems such as soft robots.
Here, we leverage the imbalance in complexity of the dynamics to learn more sample-efficiently.
We (i) abstract the task into distinct components, (ii) off-load the simple dynamics parts into the simulation, and (iii) multiply these virtual parts to generate more data \textit{in hindsight}.
Our new method, Hindsight States~(HiS), uses this data and selects the most useful transitions for training.
It can be used with an arbitrary off-policy algorithm. 
We validate our method on several challenging simulated tasks and demonstrate that it improves learning both alone and when combined with an existing hindsight algorithm, Hindsight Experience Replay~(HER).
Finally, we evaluate HiS on a physical system and show that it boosts performance on a complex table tennis task with a muscular robot.
Videos and code of the
experiments can be found on \link{http://webdav.tuebingen.mpg.de/his/}{webdav.tuebingen.mpg.de/his/}.

%% file: 1_intro.tex
Reinforcement learning~(RL) holds great potential for devising optimal behavior for challenging robotics tasks.
Difficulties in these tasks often arise from contact in object manipulation, the requirement to perform fast but accurate motions, or from hard-to-control systems like soft robots. 
While it has been shown that RL can handle such problems, the major drawback of RL methods to require large amounts of interactions with the environment remains.
This fact makes learning on real robots challenging.

Training in simulation and transferring the resulting policy to the real system is one approach to alleviate this challenge.
However, even small inaccuracies in the simulation can cause the simulated and real system to behave differently, a problem known as \emph{reality gap}.
This problem is worse for complex systems like soft robots, which are hard to model accurately.
Sim-to-real techniques attempt to mitigate the impact of the reality gap.
Many popular techniques accomplish more effective policy transfer by making the learned policies more robust to variations in the task dynamics through domain randomization~\cite{muratore_robot_2022,zhu_survey_2021}.
However, these methods are generally computationally expensive and require precise fine-tuning of the parameters, such as the amount of randomization on the dynamics parameters.
These downsides of sim-to-real approaches accumulate as tasks advance in complexity, thus, necessitating learning many complex tasks partly or completely on the real system.

Hybrid sim and real~(HySR)~\cite{buchler_learning_2022} is a way to ease the training of complex tasks on real systems.
The method is based on the insight that for some tasks, certain parts of the environment have simpler dynamics than others.
For instance, many robotic ball games consist of a robot, whose dynamics are more complex compared to those of the ball.
The mismatch in modeling complexity is exacerbated if the robot is equipped with soft components, which render modeling even more difficult.
The idea behind HySR is to keep the complicated parts of the task real, whereas the simpler parts are simulated.
This strategy yields significant practical benefits, while facilitating the transfer to the entirely real system.
First, the approach simplifies setting up the task since fewer real objects have to be handled that, for instance, are subject to wear-and-tear or require safety considerations.
Second, it can greatly simplify the reset mechanism in episodic RL tasks. 
Robots are actuated and can typically reset themselves, while environment resets often require manually moving all objects to their initial positions.
Third, data-augmentation techniques can be added to the virtual objects that would not be feasible on a fully real system.
Lastly, the simulation provides information about, e.g., the exact positions of virtual objects or about contacts that occur, which might not be easily attainable for real objects. 
This data can be used to generate labels or rewards that can aid the training process.
While HySR can simplify practical aspects of the training, the large number of real robot interactions required for training can still cause problems for real-world RL.

In this work, we present an approach that can significantly reduce the number of interactions with the real system required for learning.
The key idea behind our method, Hindsight States~(HiS), is to pair the data of a single real instance with additional data generated by concurrently simulating multiple distinct instances of the virtual part.
In the example of robot ball games, our method simulates the effect of the robot's actions on several virtual balls simultaneously.
We relabel the virtual part of each roll-out with this additional virtual data \emph{in hindsight}.
Intuitively, the agent experiences what would have happened if it had encountered a different virtual part but applied the same actions as in the original episode.
This relabeling process enables the RL agent to generate extra experience without collecting further transitions on the real system.
The hindsight data can then be used in addition to the regular training data by any off-policy RL algorithm, such as Soft Actor-Critic~(SAC)~\cite{haarnoja_soft_2018}. 
Particularly for tasks with sparse rewards, the additional data is beneficial since it increases the probability that the agent experiences positive rewards early in the training.
The contributions of this paper are the following:
\begin{enumerate}[leftmargin=.2cm]
    \item \textit{Extending HySR for sample efficient training}\\
        Our main contribution is to devise a sample efficient way to train in the Hybrid Sim and Real~(HySR) setting.
        To that end, we formalize the HySR training and, based on it, develop a novel algorithm, called Hindsight States~(HiS).
    \item \textit{Evaluation of the interplay between HER and HiS}\\
        Our experiments demonstrate that the combination of HiS and HER achieves higher sample efficiency than each method by itself. We argue that HiS and HER improve task performance in distinct and complementary manners.
    \item \textit{Thorough experimental validation of HiS on a variety of tasks}\\
        We show improved performance of HiS on the original HySR real robot table tennis task.
        Additionally, we investigate to what extent HiS can be applied to manipulation tasks and report more efficient training in these regimes~(see \fig{fig:fetch_tasks} for an overview of all experiments).
        Another finding is that HiS allows for more efficient training in entirely simulated environments w.r.t. wall-clock time.  
\end{enumerate}

%% file: 2_related_work.tex
Our new method, HiS, shares some aspects with and takes inspiration from other works. 
In the following section, we discuss these similarities.

Hindsight Experience Replay~(HER)~\cite{andrychowicz_hindsight_2017} improves sample efficiency for sparse goal-conditioned RL tasks by relabeling the goal states of transitions in hindsight.
The method replaces the original goal with a state reached by the agent.
Then it trains an off-policy RL agent on a combination of original and relabeled data.
This way, the agent receives additional positive feedback for reaching the relabeled goal, which can be more instructive than the negative feedback for missing the original goal.
\citet{rauber_hindsight_2019} extend the idea to on-policy algorithms through the use of importance sampling.
Dynamic Hindsight Experience Replay~(DHER)~\cite{fang_dher_2018} is a version of HER that supports dynamic goals, which change during the episode.
The method makes the idea of relabeled goals applicable to tasks like grasping moving objects.
While HER samples hindsight goals uniformly, recent methods prioritize goals based on how instructive the resulting transitions are for the agent.
These approaches sample hindsight goals that guide the agent toward the true goal~\cite{bai_guided_2019,ren_exploration_2019,fang_curriculum_2019} or result in a large temporal difference~(TD) error~\cite{liu_generating_2020}.
Other methods sample hindsight goals uniformly and prioritize more informative transitions when sampling from the replay buffer, akin to PER~\cite{schaul_prioritized_2015}.
\citet{zhao_energy_2018} prioritize trajectories that demonstrate difficult behavior.
They quantify difficulty by the increase of the system's energy over the course of the trajectory.
\citet{beyene_prioritized_2022} primarily sample hindsight transitions that incur a large TD error.
Similar to HER and its extensions, our method also relabels in hindsight.
However, it focuses on exchanging the virtual part of the state rather than the goal.

HiS and sim-to-real techniques utilize simulations to devise a policy that transfers to reality. Different techniques, such as domain randomization~\cite{muratore_robot_2022,peng_sim--real_2018,mahler_learning_2019} and domain adaptation~\cite{james_sim_2019,higgins_darla_2017,zhang_vr_2019} were proposed to mitigate the impact of the reality gap.
HiS, on the other hand, assumes the reality gap to be sufficiently small for certain components of the environment and devises a way to be more efficient in the hybrid sim and real setting.
Sim-to-real methods could be used on top of HiS to close the remaining gap.

Other works have investigated leveraging real and simulated data of the same task to improve learning performance.
For instance, \citet{kang_generalization_2019} learn a perception system using simulated and a reward model based on real data.
They use model predictive control with the learned reward model to solve quadrotor navigation tasks.
\citet{dicastro_sim_2021} combine cheap but imprecise simulated with expensive real transitions to learn a policy for the real task.
HiS, in contrast, investigates more efficient training by mixing simulated and real components in each transition.

Utilizing offline RL~\cite{lange_batch_2012,levine_offline_2020} can also alleviate some of the problems related to training on real systems by making use of existing datasets.
The central issue of offline RL is the distribution shift between the behavior policy, which collected the dataset, and the trained policy~\cite{levine_offline_2020}.
Recent works constrain the trained policy to be similar to the behavior policy~\cite{wu_behavior_2019,siegel_keep_2020,kumar_stabilizing_2019}, modify the objective of the Q-function to be conservative with respect to samples not in the dataset~\cite{kumar_conservative_2020,yu_combo_2021}, or penalize uncertainty in rollouts of learned models \cite{yu_mopo_2020,rafailov_offline_2021}.

%% file: 3_method.tex
The core components of HiS comprise the generation of parallel virtual trajectories, as well as the criteria to select among the additional transitions generated with HiS.
This section introduces each of these components, as well as the HySR setup that HiS builds upon.

\begin{algorithm}[t]
    \caption{Hindsight States~(HiS)}
	\label{alg:his}
	 \begin{algorithmic}[1]
		\State Initialize off-policy RL algorithm $\mathbb{A}$, replay buffer $R$, choose criterion $c$
		\For{episode $i =1,2,\ldots$}
			\State Sample $\sv{1:T} \sim \Drec{}$ and initialize real system to $\sr{1}$
			\For{$t=1, \ldots , T$} 
				\State Sample action $\act{t}$ from $\mathbb{A}$ given $\s{t} = [\sr{t}, \sv{t}]$
				\State Apply $\act{t}$ to the real system and observe $\sr{t+1}$
				\State $\sv{t+1} \sim \texttt{sim}([\sr{t}, \sv{t}], \act{t})$ \textrm{  or  }  take $\sv{t+1}$ from $\sv{1:T}$
			\EndFor
			\For{$t=1, \ldots , T$}  \Comment{standard replay}
				\State Store $([\sr{t}, \sv{t}], \act{t}, r([\sr{t}, \sv{t}], \act{t}), [\sr{t+1}, \sv{t+1}]$) in $R$
			\EndFor
			\State Initialize temporary buffer $R_{\textrm{temp}}$  \Comment{HiS replay}
			\For{$j = 1, 2, \ldots$}
				\State Sample $\svi{1:T}{m}{} \sim \Drec{}$
				\For{$t=1, \ldots , T$}
					\State $\svi{t+1}{m}{} \sim \texttt{sim}([\sr{t}, \svi{t}{m}{}], \act{t})$ \textrm{  or  }  take $\svi{t+1}{m}{}$ from $\svi{1:T}{m}{}$
                    \State $\tri{t}{m} = ([\sr{t}, \svi{t}{m}{}], \act{t}, r([\sr{t}, \svi{t}{m}{}], \act{t}), [\sr{t+1}, \svi{t+1}{m}{}])$
					\State Store relabeled transition $\tri{t}{m}$ in $R_{\textrm{temp}}$
				\EndFor
			\EndFor
			\State Every few episodes:
            \Statex\hspace{\algorithmicindent}\hspace{0.5cm}Add transitions from $R_{\textrm{temp}}$ to $R$ using criterion $c$
			\State Every few steps:
            \Statex\hspace{\algorithmicindent}\hspace{0.5cm}Perform optimization on $\mathbb{A}$ with replay buffer $R$ 
		\EndFor
	 \end{algorithmic}
\end{algorithm}

\subsection{RL Preliminaries}
We consider tasks formulated as a discounted Markov decision process, which is defined by the tuple $\mathcal{M} = (\mathcal{S}, \mathcal{A}, \mathcal{P}, \gamma, r)$, where $\mathcal{S}$ denotes the state space and $\mathcal{A}$ the action space.
The transition dynamics $\mathcal{P}(\s{t + 1} | \s{t}, \act{t})$ represent the probability of reaching state $\s{t + 1} \in \mathcal{S}$ after executing action $\act{t} \in \mathcal{A}$ in state $\s{t} \in \mathcal{S}$.
Furthermore, the agent receives a scalar reward $\rew{t}=r(\s{t},\act{t})$ at each time step. 
The goal of RL is to find an optimal policy $\pi^{\star}$, which maximizes the discounted total return 
\begin{align}
    R =  \EX_\pi \left[\sum_{t}^{T} \gamma^{\,t} \rew{t}\right]\,,
    \label{eq:return}
\end{align} 
where $T$ is the time horizon and $\gamma \in [0, 1)$ is the discount factor.
The agent collects transitions $\tr{t}=(\s{t},\act{t},\rew{t},\s{t+1})$ by interacting with the environment and uses them to iteratively update its policy $\pi$. 
Many popular off-policy RL algorithms, such as SAC~\cite{haarnoja_soft_2018} or Deep Q-Networks~(DQN)~\cite{mnih_playing_2013}, store these transitions in a buffer~$\D{}$~\cite{lin_self-improving_1992} and replay them when updating, for instance, the action value function $Q(\s{t},\act{t})$.
This replay buffer is fixed-sized and contains the most recent transitions.
A ring buffer, which replaces the oldest with the most recent transition, is a popular implementation. 
The $Q$-function update involves minimizing a variant of the TD error
\begin{align}
    \delta(\s{},\act{},\rew{},\s{}')=\rew{}+\gamma\,\max_{\act{}'\in\mathcal{A}} Q(\s{}',\mathbf{a}')-Q(\s{},\act{})\,,
    \label{eq:td}
\end{align}
where variations might include an alternative way to select the next action $\act{}'$ or an $n$-step TD error formulation $\delta_{n}=\sum_{k=0}^{n-1}\gamma^k\rew{t+k}+\gamma^{n}\max_{\act{}'}Q(\s{t+n},\act{}')-Q(\s{t},\act{t})$. 
The TD error represents a measure of surprise: the higher the absolute value of $\delta$, the more does this particular transition influence the update of the $Q$-function.
For this reason, the TD error metric is commonly used to establish a ranking among transitions, for example in prioritized sweeping~\cite{moore_prioritized_1993,peng_efficient_1993} and prioritized experience replay~(PER)~\cite{schaul_prioritized_2015}.
The $Q$-function loss employs the replay buffer $\D{}$ and the squared TD error
\begin{align}
    J(\theta)=\EX_{(\s{},\act{},\rew{},\s{}')\sim\D{}}\left[(\rew{}+\gamma\,\max_{\act{}'\in\mathcal{A}} Q_{\theta^\textrm{old}}(\s{}',\act{}')-Q_\theta(\s{},\act{}))^2\right]\,,
\end{align}
where the old parameters $\theta^\textrm{old}$ are kept fixed during optimization of the current policy parameters $\theta$ and transitions are sampled from $\D{}$ according to a distribution that traditionally is uniform but can vary such as in PER.  

\subsection{Hybrid Sim and Real Training}
\label{subsec:HySR}
Hybrid Sim and Real~(HySR) training~\cite{buchler_learning_2022} is the foundation for Hindsight States~(HiS).
Intuitively, the idea of HySR is to gain practical benefits by offloading the part of the task that is easier to model to the simulation.
At the same time, the difficult part is kept real during training.
More formally, HySR assumes that the Markovian state $\s{}$ splits into a real state $\sr{}$ and a virtual state $\sv{}$
\begin{align}
	\s{} = [\sr{}, \sv{}]\,
	\label{eq:state}
\end{align}
and that the dynamics governing the real part $p(\sr{t+1}|[\sr{t},\sv{t}],\act{t})$ are more complex than those governing the virtual part $p(\sv{t+1}|[\sr{t},\sv{t}],\act{t})$.
 To better transfer the learned policy after HySR training to the fully real setup, HySR keeps the complicated dynamics real and expects the virtual dynamics to be sufficiently accurate.
Another requirement is that the virtual state does not influence the real state
\begin{align}
	p(\sr{t+1}|[\sr{t},\sv{t}],\act{t})=p(\sr{t+1}|\sr{t},\act{t})
\end{align}
since mapping forces from the virtual to the real part can be intricate, especially onto complex or unknown dynamics.

The virtual dynamics do not necessarily simulate the part for the full duration of the training.
Before the first contact between real and virtual part, HySR replays recorded data of the virtual part instead of simulating it to minimize the accumulation of model error.
To that end, HySR first uniformly samples a virtual trajectory $\taurec{}$ from a database $\Drec$ containing $\Nr$ recorded trajectories.
\begin{equation}
    \Drec=\left[\svi{1}{n}{\textrm{rec}},\ldots,\svi{T}{n}{\textrm{rec}}\right]\!\strut_{n=1}^{\Nr}
\end{equation}
After contact of the virtual and the real part, the simulated dynamics determine the consecutive motion, as no information about the latter part can be deduced from the recordings.
A complete HySR training trajectory,
\begin{align}
    \tauhysr{}=[\tr{1}^\textrm{rec},\ldots,\tr{t_\textrm{c}}^\textrm{rec},\tr{t_\textrm{c}+1}^\textrm{sim},\dots,\tr{T}^\textrm{sim}]\,,
    \label{eq:HySR_traj}
\end{align}
hence, consists of both transitions replayed from the database $\tr{t}^\textrm{rec}=\left([\sr{t},\svi{t}{}{\textrm{rec}}],\act{t},\rew{}([\sr{t},\svi{t}{}{\textrm{rec}}],\act{t}),[\sr{t+1},\svi{t+1}{}{\textrm{rec}}]\right)$ and simulated transitions $\tr{t}^\textrm{sim}=\left([\sr{t},\svi{t}{}{\textrm{sim}}],\act{t},\rew{}([\sr{t},\svi{t}{}{\textrm{sim}}],\act{t}),[\sr{t+1},\svi{t+1}{}{\textrm{sim}}]\right)$, assuming that the contact happened at time $t_\textrm{c}$.
During training, the policy samples actions conditioned on the complete state from \eq{eq:state}, in which the information of whether the virtual parts are replayed or simulated is not included.

\begin{figure*}
    \centering
    \subfloat[MDP]{
      \centering%
	     \input{figures/tikz_figs/MDP_HER_graph_.tikz}
  	}
    \hspace{0.5cm}
    \subfloat[HER]{
      \centering%
	   \input{figures/tikz_figs/HER.tikz}
	}
    \hspace{0.5cm}
    \subfloat[HySR]{
	     \centering%
	   \input{figures/tikz_figs/HySR.tikz}
	  }
    \hspace{0.5cm}
    \subfloat[HiS]{
	   \centering%
	   \input{figures/tikz_figs/HIS.tikz}
   }
    \caption{
	       Graphical models of a general MDP, HER, HySR, and HiS.
	       All blue and red elements indicate a difference to the general MDP.
	       Red indicates components that are relabeled in hindsight.
	       HER extends the MDP with a goal state and relabels goals.
	       HySR separates the state into a virtual and a real part. The additional arrows indicate the dynamics.
	       HiS combines key ideas from HySR and HER that are state separation and dynamics as well as relabeling, respectively.
	    }
    \label{fig:MHHH}
\end{figure*}
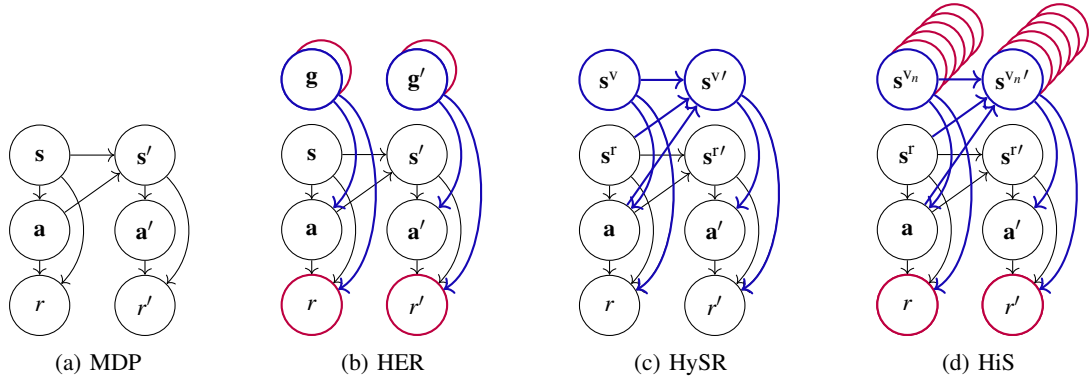

HySR is particularly suitable for applications where 
(i) the complexity of the virtual and real dynamics is unbalanced,
(ii) training with the real instance of the virtual part is handy or safer,
(iii) contact of the separated parts occurs relatively far into each roll-out to make better use of the recorded data,
(iv) and the influence that the virtual part has on the real part is negligible and the effect in the opposite direction is known and transfers well to the real world.
All these points apply to many ball games, such as football, cricket, baseball, or basketball, where an object has to reach a goal state through an interaction with the player.
Although, the efficacy of HySR has only been shown for robot table tennis so far, training in the HySR setting is not limited to ball games. 
For instance, for tasks like slide-pushing or throwing objects (in contrast to picking them up), we could combine objects of different weights or materials with a single arm motion.  
Points (i) to (iv) also apply to tasks such as avoiding moving obstacles since contact should be avoided in the first place.
For example, a robot waking in a cluttered and dynamic environment or an autonomous car driving through a crowded city.

\subsection{Hindsight States}
\label{sec:his}

HySR also enables the generation of additional data, and \textit{Hindsight States}~(HiS) is our proposed way to leverage them for efficient training.
HiS implements the idea that we can generate additional virtual data in the HySR setting. Specifically, it generates trajectories $\tauhis{} = [\tr{1}^m, \ldots, \tr{T}^m]$ with $m = 1, \ldots, M$ for each HySR trajectory $\tauhysr{} = [\tr{1}, \ldots, \tr{T}]$.
For each of these $M$ trajectories, we sample a different series of virtual states $\svi{1:T}{m}{} \sim \Drec{}$.
We augment the HySR trajectory to reflect what would have happened if the virtual states were $\svi{1:T}{m}{}$.
In particular, we generate the hindsight transitions 
\begin{equation}
    \tri{t}{m} = \left([\sr{t},\svi{t}{m}{}],\act{t},\rew{}([\sr{t},\svi{t}{m}{}],\act{t}),[\sr{t+1},\svi{t+1}{m}{}]\right)
\end{equation}
for each transition $\tr{t} = \left([\sr{t},\sv{t}],\act{t},\rew{}([\sr{t},\sv{t}],\act{t}),[\sr{t+1},\sv{t+1}]\right)$ generated by HySR.
The real states $\sr{t}, \sr{t+1}$ and the action $\act{t}$ remain the same as in the HySR transition, but the virtual states are relabeled in hindsight with the different instances of the virtual states $\svi{t}{m}{}, \svi{t+1}{m}{}$, which are either taken from the pre-recorded or simulated data.
The reward of the hindsight transitions is then calculated according to the relabeled state.
Note that in this manner, HiS generates a wide bouquet of how the real part could have interacted with instances of the virtual object.
For this reason, HiS provides valuable feedback on the quality of this particular action sequence in different situations.
In the table tennis example depicted in \fig{sfig:image_table_tennis_sim}, the racket hits a whole cloud of balls with different speeds and trajectories; hence, the resulting balls arrive at very distinct landing points.

\paragraph{Picking Strategy}
The naive approach in which all HiS trajectories are added to a replay buffer and sampled by an off-policy algorithm can be detrimental to the learning process.
A critical influence on the sample complexity of RL algorithms is the degree of \textit{off-policyness} of transitions in the replay buffer~\cite{fedus_revisiting_2020}. 
Off-policy data are challenging because they have been collected under multiple dissimilar policies, but the expectation in the return in \eq{eq:return} is computed w.r.t.\ the distribution induced by the current policy $\pi$.
This distributional shift adds bias and variance to the estimation of the return and leads to incorrect estimations of the $Q$ values.
Typically, the size of the buffer determines the degree of off-policyness of the buffer since older transitions are more likely to stem from older and hence dissimilar policies~\cite{fedus_revisiting_2020}.
The additional HiS trajectories add to the off-policyness of the buffer since the actions in the HiS trajectories were generated conditioned on the original HySR states instead of the relabeled states.

For this reason, HiS training includes a selection strategy that chooses the transitions that accelerate the learning process most.
We propose to rank HiS transitions based on the reward $\rew{}(\s{},\act{})$ or the TD error $\delta(\s{},\act{},r,\s{}')$ from \eq{eq:td}.
The reasoning behind prioritizing higher rewards is to experience high rewards early on, which is especially useful in a sparse reward setting.
The TD error corresponds to the amount of influence on the $Q$-function update.
Hence, it is a sensible choice for ranking transitions.
The selection process is based on the absolute value of the TD error $|\delta(\s{},\act{},r,\s{}')|$ since updating with both high and low-value transition ascribes to bringing the action value function closer to its optimal version $Q^\ast$.
It is instrumental how these measures~(TD error or reward) are applied to the transitions of the HiS trajectories. 
We study two ways: (i) applying the measure on each transition and (ii) on the whole set of transitions of a trajectory by taking the sum of the reward or TD error over the whole trajectory.
The first implementation represents the straightforward approach, whereas the second way tests the hypothesis that all transitions of a trajectory with high value of the measure are significant.
The idea of the latter point has been explored in the context of prioritized replay, where it turned out that transitions correlating in time have correlating TD errors~\cite{brittain_prioritized_2019}. 

\begin{figure}[t]
	\centering
	\vspace{-.25cm}
	\subfloat[simulated robot]{
		\centering\scriptsize%
		\hspace{-.25cm}
		\input{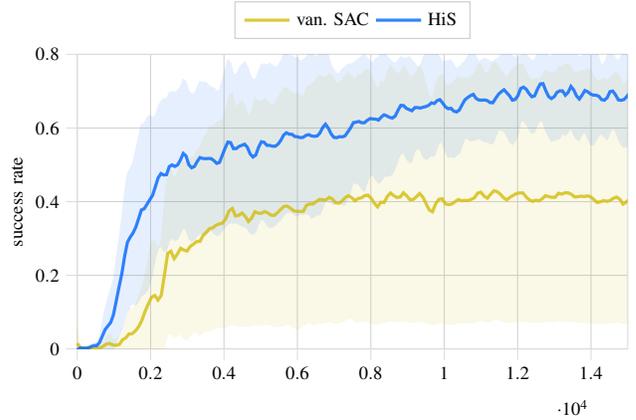}
		\label{sfig:learning_curve_tt_sim}
	} \\
   \vspace{0.5cm}
	\subfloat[real robot]{
		\centering\scriptsize%
		\hspace{-.25cm}
		\input{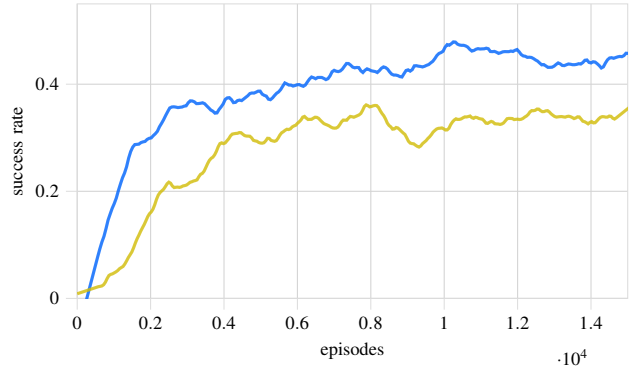}
		\label{sfig:learning_curve_tt_real}
	}
	\caption{Results on the table tennis task. HiS increases sample-efficiency and asymptotic performance with respect to the number of robot steps.}
    \label{fig:learning_curve_tt}
\end{figure}

Having defined the criteria to rank HiS transitions, we now present the methodology for determining the number of HiS transitions to be added to the replay buffer.
For inclusion into the buffer, criterion $c$ has to both (i) exceed a threshold $\psi_c$ and (ii) be among the $k_c$ best transitions according to the criterion. 
Case (i) prevents transitions from being added that are among the best of their peers for this round but too low in value. 
Condition (ii) controls the maximum amount of off-policy data in the replay buffer. 
\alg{alg:his} summarizes HiS in detail. 
\paragraph{Relation to HER}
HiS shares some aspects with HER~\cite{andrychowicz_hindsight_2017}.
HER trains a goal-conditioned policy and relabels the desired goal with the actually achieved goal in hindsight.
Both HER and HiS use hindsight data to learn more sample efficiently, and are especially useful in a sparse reward setting.
In contrast, HiS relabels virtual states instead of goals.

\fig{fig:MHHH} depicts the differences between a standard MDP, HySR, HER, and HiS with respect to their corresponding graphical models.
Note that the HER setting also allows for multiple goals in each transition and \cite{fang_dher_2018} introduced dynamics into the goal space of HER.
As can be seen in \fig{fig:MHHH}, goals and dynamic goals \cite{fang_dher_2018} can be seen as special cases of virtual states. Therefore, our method can also be applied to the goal-conditional setting.
The difference to HER, when applied to this setting, is that HER will sample goals such that they correspond to specific achieved goals (e.g., the goal sampled in hindsight is the final position of the robot gripper). 
HiS, however, will sample from the initial goal distribution and then use a prioritization strategy to select from the set of sampled goals.
With enough goals sampled and a corresponding prioritization strategy, HiS could converge toward doing the same as HER (e.g., HiS samples enough goals, such that the prioritization strategy can pick goals that are similar to what the HER strategy would have chosen). 
In this setting, the additional complexity of HiS compared to HER seems unnecessary. 
However, in the  general HySR setting, HER cannot be applied to virtual states, while the application of HiS is possible.

On a task, that has both a goal state and a virtual state, HER can be applied on top of HiS.
\paragraph{Combining HiS and HER}
HER and HiS are addressing the problem of sparsity during learning from two different directions. 
HER generates additional experiences that can be beneficial to the learning process because they solve the goal and have high rewards, but because only goals are relabeled, HER does not directly help exploring the state space.
On the other hand, by creating additional virtual states of the environment, HiS helps exploring the state space.
Contrary to HER, experiences created by HiS might not reach the goal or be of high reward.
To get some of the benefits of HER in a goal-directed HySR setting, HiS could sample numerous hindsight trajectories until some of these trajectories also reach the goal.
For a challenging and sparse task, however, this strategy might be computationally expensive or even infeasible.

Instead, a combination of HER and HiS can be beneficial for such tasks, and utilize the strengths of both.
Practically, the combination can be implemented by following Algorithm \ref{alg:his}, and adding complete HiS episodes into the replay buffer that is used by HER. HER relabels the goal of the HiS trajectories, and the resulting trajectories with relabeled virtual state and relabeled goal can then be sampled to optimize the policy.

%% file: figures/tikz_figs/MDP_HER_graph_.tikz
\begin{tikzpicture}[every node/.style={font=\small}]
	\begin{pgfonlayer}{nodelayer}
		\node [style=circle standard] (4) at (-2, 0) {$\mathbf{s}$};
		\node [style=circle standard] (5) at (0.8, 0) {$\mathbf{s'}$};
		\node [style=circle standard] (6) at (-2, -2) {$\mathbf{a}$};
		\node [style=circle standard] (7) at (0.8, -2) {$\mathbf{a'}$};
		\node [style=circle standard] (8) at (-2, -4) {$r$};
		\node [style=circle standard] (9) at (0.8, -4) {$r'$};
	\end{pgfonlayer}
	\begin{pgfonlayer}{edgelayer}
		\draw [style=arrow] (4) to (5);
		\draw [style=arrow] (6) to (8);
		\draw [style=arrow] (4) to (6);
		\draw [style=arrow] (5) to (7);
		\draw [style=arrow] (7) to (9);
		\draw [style=arrow, bend left=45] (4) to (8);
		\draw [style=arrow, bend left=45] (5) to (9);
		\draw [style=arrow] (6) to (5);
	\end{pgfonlayer}
\end{tikzpicture}

%% file: figures/tikz_figs/HER.tikz
\begin{tikzpicture}[every node/.style={font=\small}]
	\begin{pgfonlayer}{nodelayer}
		\node [style=red circle, minimum size=0.79cm] (34) at (14.25, 2) {};
		\node [style=red circle, minimum size=0.79cm] (35) at (17.05, 2) {};
		\node [style=circle standard] (38) at (14, -0.25) {$\mathbf{s}$};
		\node [style=circle standard] (39) at (16.8, -0.25) {$\mathbf{s'}$};
		\node [style=circle standard] (40) at (14, -2.25) {$\mathbf{a}$};
		\node [style=circle standard] (41) at (16.8, -2.25) {$\mathbf{a'}$};
		\node [style=circle standard] (42) at (14, -4.25) {$r$};
		\node [style=circle standard] (43) at (16.8, -4.25) {$r'$};
		\node [style=red circle] (48) at (14, -4.25) {$r$};
		\node [style=red circle] (49) at (16.8, -4.25) {$r'$};
		\node [style=blue circle] (50) at (14, 1.75) {};
		\node [style=blue circle] (51) at (14, 1.75) {$\mathbf{g}$};
		\node [style=blue circle] (52) at (16.8, 1.75) {};
		\node [style=blue circle] (53) at (16.8, 1.75) {$\mathbf{g'}$};
	\end{pgfonlayer}
	\begin{pgfonlayer}{edgelayer}
		\draw [style=arrow] (38) to (39);
		\draw [style=arrow] (40) to (42);
		\draw [style=arrow] (38) to (40);
		\draw [style=arrow] (39) to (41);
		\draw [style=arrow] (41) to (43);
		\draw [style=arrow, bend left=45] (38) to (42);
		\draw [style=arrow, bend left=45] (39) to (43);
		\draw [style=arrow] (40) to (39);
		\draw [style=blue arrow, bend left=45] (51) to (40);
		\draw [style=blue arrow, bend left=45] (53) to (41);
		\draw [style=blue arrow, bend left=60, looseness=0.75] (51) to (42);
		\draw [style=blue arrow, bend left=60, looseness=0.75] (53) to (43);
	\end{pgfonlayer}
\end{tikzpicture}

%% file: figures/tikz_figs/HYSR.tikz
\begin{tikzpicture}[every node/.style={font=\small}]
	\begin{pgfonlayer}{nodelayer}
		\node [style=circle standard] (2) at (10, 0) {$\mathbf{s}^\textrm{r}$};
		\node [style=circle standard] (3) at (12.8, 0) {$\mathbf{s}^\textrm{r}$$'$};
		\node [style=circle standard] (4) at (10, -2) {$\mathbf{a}$};
		\node [style=circle standard] (5) at (12.8, -2) {$\mathbf{a'}$};
		\node [style=circle standard] (6) at (10, -4) {$r$};
		\node [style=circle standard] (7) at (12.8, -4) {$r'$};
		\node [style=blue circle] (15) at (10, 2) {$\mathbf{s}^{\textrm{v}}$};
		\node [style=blue circle] (16) at (12.8, 2) {$\mathbf{s}^{\textrm{v}}$$'$};
	\end{pgfonlayer}
	\begin{pgfonlayer}{edgelayer}
		\draw [style=arrow] (2) to (3);
		\draw [style=arrow] (4) to (6);
		\draw [style=arrow] (2) to (4);
		\draw [style=arrow] (3) to (5);
		\draw [style=arrow] (5) to (7);
		\draw [style=arrow, bend left=45] (2) to (6);
		\draw [style=arrow, bend left=45] (3) to (7);
		\draw [style=arrow] (4) to (3);
		\draw [style=blue arrow] (15) to (16);
		\draw [style=blue arrow] (2) to (16);
		\draw [style=blue arrow, bend left=45] (15) to (4);
		\draw [style=blue arrow, bend left=45] (16) to (5);
		\draw [style=arrow, bend left=60, looseness=0.75] (15) to (6);
		\draw [style=arrow, bend left=60, looseness=0.75] (16) to (7);
		\draw [style=blue arrow, bend left=60, looseness=0.75] (15) to (6);
		\draw [style=blue arrow, bend left=60, looseness=0.75] (16) to (7);
		\draw [style=blue arrow] (4) to (16);
	\end{pgfonlayer}
\end{tikzpicture}

%% file: figures/tikz_figs/HIS.tikz
\begin{tikzpicture}[every node/.style={font=\small}]
	\begin{pgfonlayer}{nodelayer}
		\node [style=red circle, minimum size=0.75cm] (86) at (20.05, 2.5) {};
		\node [style=red circle, minimum size=0.76cm] (87) at (19.8, 2.25) {};
		\node [style=red circle, minimum size=0.77cm] (88) at (19.55, 2) {};
		\node [style=red circle, minimum size=0.78cm] (89) at (19.3, 1.75) {};
		\node [style=red circle, minimum size=0.79cm] (90) at (19.05, 1.5) {};
		\node [style=red circle, minimum size=0.75cm] (110) at (17.25, 2.5) {};
		\node [style=red circle, minimum size=0.76cm] (111) at (17, 2.25) {};
		\node [style=red circle, minimum size=0.77cm] (112) at (16.75, 2) {};
		\node [style=red circle, minimum size=0.78cm] (113) at (16.5, 1.75) {};
		\node [style=red circle, minimum size=0.79cm] (114) at (16.25, 1.5) {};
		\node [style=circle standard] (117) at (16, -0.75) {$\mathbf{s}^\textrm{r}$};
		\node [style=circle standard] (118) at (18.8, -0.75) {$\mathbf{s}^\textrm{r}$$'$};
		\node [style=circle standard] (119) at (16, -2.75) {$\mathbf{a}$};
		\node [style=circle standard] (120) at (18.8, -2.75) {$\mathbf{a'}$};
		\node [style=circle standard] (121) at (16, -4.75) {$r$};
		\node [style=circle standard] (122) at (18.8, -4.75) {$r'$};
		\node [style=blue circle] (125) at (16, 1.25) {$\mathbf{s}^{\textrm{v}_n}$};
		\node [style=blue circle] (126) at (18.8, 1.25) {$\mathbf{s}^{\textrm{v}_n}$$'$};
		\node [style=red circle] (130) at (16, -4.75) {};
		\node [style=red circle] (131) at (16, -4.75) {$r$};
		\node [style=red circle] (132) at (18.8, -4.75) {};
		\node [style=red circle] (133) at (18.8, -4.75) {$r'$};
	\end{pgfonlayer}
	\begin{pgfonlayer}{edgelayer}
		\draw [style=arrow] (117) to (118);
		\draw [style=arrow] (119) to (121);
		\draw [style=arrow] (117) to (119);
		\draw [style=arrow] (118) to (120);
		\draw [style=arrow] (120) to (122);
		\draw [style=arrow, bend left=45] (117) to (121);
		\draw [style=arrow, bend left=45] (118) to (122);
		\draw [style=arrow] (119) to (118);
		\draw [style=blue arrow] (125) to (126);
		\draw [style=blue arrow] (117) to (126);
		\draw [style=blue arrow, bend left=45] (125) to (119);
		\draw [style=blue arrow, bend left=45] (126) to (120);
		\draw [style=blue arrow, bend left=60, looseness=0.75] (125) to (121);
		\draw [style=blue arrow, bend left=60, looseness=0.75] (126) to (122);
		\draw [style=blue arrow] (119) to (126);
	\end{pgfonlayer}
\end{tikzpicture}

%% file: 4_experiments_new.tex
Our new method HiS promises improved sample efficiency in the HySR setting.
The evaluation of HiS is carried out in two regimes: We run extensive experiments on 1) several simulated tasks, where the HySR assumptions apply, and 2) a challenging real robot table tennis task. 
Furthermore, we investigate to what extent HiS is useful in manipulation tasks and illustrate the efficiency of the combination of HER and HiS. 

\subsection{Applying HiS to the Original HySR Table Tennis Task}
The work that originally introduced HySR~\cite{buchler_learning_2022} learned to play table tennis with a muscular robot and naturally separated between the hard-to-control soft muscular robot~\cite{buchler_lightweight_2016, buchler_learning_2023} and the relatively simple ball model.
Different from the original task, we refrain from using the hand-crafted dense reward function and use a simple sparse reward that returns one in case the ball lands within a circle of radius \SI{40}{\cm} and zero otherwise. 
To apply HiS, we replay 20 parallel instances of the virtual ball sampled from a set of 100 recorded ball trajectories. 
We apply HiS on top of SAC~\cite{haarnoja_soft_2018} and compare with vanilla SAC as a baseline. 
For trajectory selection, we use the reward criterion with threshold $\psi_c = 0.5$. 
The~\nameref{sec:appendix} contains an in-depth description of the experiment details and the parameters of each experiment. It also contains an evaluation of different selection criteria.

The experiments depicted in~\fig{fig:learning_curve_tt} demonstrate that HiS enables us to learn this task with fewer robot time steps in both, the fully simulated and the HySR setting. 
The simulated experiments from~\fig{sfig:learning_curve_tt_sim} show the mean and variance of ten runs with different random seeds. 
HiS achieves a maximum average performance of appr.\ 70\% success rate, whereas vanilla SAC reaches appr.\ 40\%.
HiS matches the asymptotic performance of SAC with appr.\ 29\% of the samples that SAC needs to reach its asymptotic performance~(2k and 7k episodes).
Another observation is that the variance between different runs is smaller compared to SAC.

HiS is also more efficient when using the real robot. 
\fig{sfig:learning_curve_tt_real} shows one training run for vanilla SAC and HiS. 
HiS achieves around 45\% success rate, while SAC reaches only appr.\ 35\%.
Similarly to the simulated experiment, HiS matches SAC's maximum success rate with appr.\ 38\% of the number of samples collected on the real robot~(3k and 8k episodes).

\begin{figure}[t]
\centering
\vspace{-.25cm}
\subfloat[\texttt{FetchPush}]{
    \centering\scriptsize%
    \hspace{-.25cm}
    \scalebox{1.0}{\input{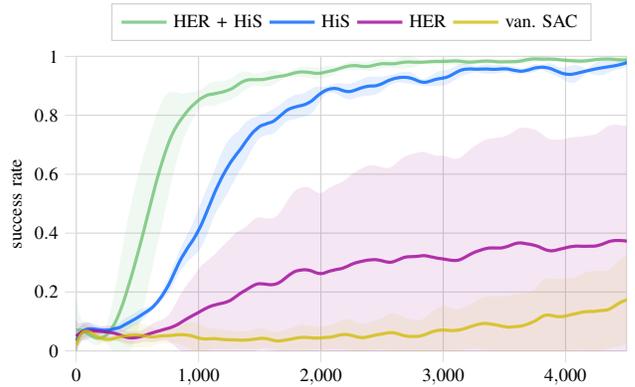}}
    \label{sfig:learning_curve_gym_push}
}\\
\vspace{0.5cm}
\subfloat[\texttt{FetchSlide}]{
    \centering\scriptsize%
    \hspace{-.25cm}
    \scalebox{1.0}{\input{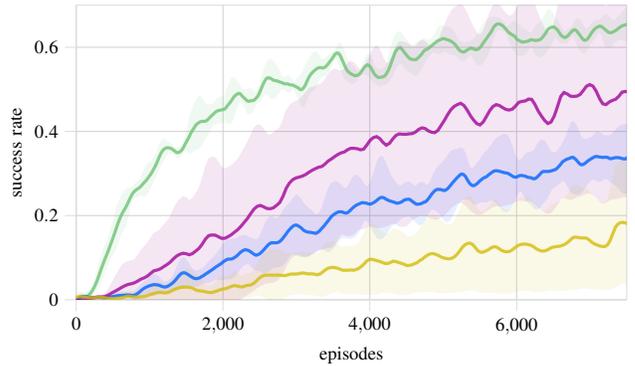}}
    \label{sfig:learning_curve_gym_slide}
}
\caption{Results on the simulated Gym robotics tasks \texttt{FetchPush} and \texttt{FetchSlide}. HER and HiS both learn faster than SAC with respect to the number of robot steps. Combining HER and HiS gives the best results.}
\label{fig:learning_curve_gym_rob}
\end{figure}

\subsection{HySR and HiS on Manipulation Tasks}

\texttt{FetchPush} and \texttt{FetchSlide} \cite{andrychowicz_hindsight_2017} are two benchmark manipulation tasks for goal-conditioned RL.
These two tasks are based on existing robot hardware and simulated using MuJoCo~\cite{todorov_mujoco:_2012}.
In the Pushing task, the goal is to move a box to a target location.
The robot's gripper is locked to prevent grasping.
The Sliding task is slightly different: Here the objective is to move a puck to a goal position that is outside the robot’s reach, thus, necessitating a sliding strategy.
Both tasks satisfy the HySR assumptions and, hence, our method could be applied to learn a policy on a real robot.
In both tasks, the manipulated object adheres to simple dynamics.
The arm is rigid and small disturbances of the contact with the object are quickly compensated by a position controller.
Thus, the influence of forces transferred from the object to the robot is small.

However, both tasks use a robot with relatively simple dynamics that can be simulated accurately.
Sim-to-real training has been shown to be successful in solving these tasks~\cite{andrychowicz_hindsight_2017} and would therefore be the better choice for this kind of system.
However, when executing similar tasks with a more complex robot, for instance one equipped with soft parts, or a less accurate position controller, the larger discrepancies between simulation and reality would likely degrade the performance of the transferred policy.
HySR would be well-suited for learning a policy on such a robot.
Our experiments show that on top of the practical benefits coming with HySR, HiS could greatly improve sample efficiency for these tasks, especially when combined with HER.
We will analyze the results in detail in the next subsection.
Furthermore, we will show, that when learning a policy in a sim-to-real fashion, HiS can still help to speed up learning during the simulation stage.

\subsection{Combining HER and HiS}

In \sect{sec:his}, we have discussed the differences and similarities between HER and HiS.
Due to the goal-conditioned nature of the \texttt{FetchPush} and \texttt{FetchSlide} tasks, they are suitable to experimentally investigate the relationship of HER and HiS.
\fig{fig:learning_curve_gym_rob} shows the results on the Pushing and Sliding tasks. 
To apply HiS, we simulate 100 parallel instances of the virtual object. 
For virtual trajectory selection, we use trajectories where the robot moves the object.
Similar to the table tennis task, we apply HiS and/or HER on top of SAC, which also serves as a baseline.

On both tasks, HER and HiS learn faster than vanilla SAC. HiS learns significantly faster than HER on the Pushing task, solving it almost perfectly in only 4k episodes, while HER surpasses HiS on the Sliding task.
The best performance, however, is achieved by HER and HiS in combination on both tasks.
Similar to the table tennis task, we also find that HiS alone as well as in combination with HER significantly reduces the variance between different runs.

\begin{figure}[t]
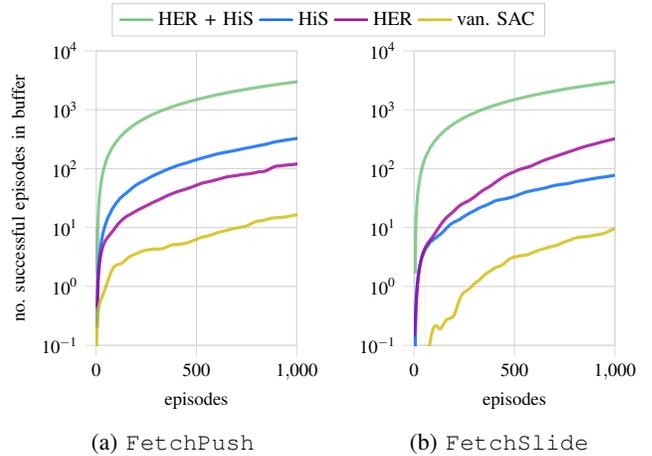

\centering
\scalebox{0.75}{\input{figures/plot_legend}}
\vspace{-.25cm}
\subfloat[\texttt{FetchPush}]{
    \centering\scriptsize%
    \hspace{-.25cm}
    \scalebox{1.0}{\input{figures/plot_gym_push_traj}}
    \label{sfig:succ_traj_push}
}
\subfloat[\texttt{FetchSlide}]{
    \centering\scriptsize%
    \hspace{-.25cm}
    \scalebox{1.0}{\input{figures/plot_gym_slide_traj}}
    \label{sfig:succ_traj_slide}
}
\vspace{0.5cm}
\caption{Number of successful trajectories added to the RL buffer within the first 1000 episodes of learning for the \texttt{FetchPush} and \texttt{FetchSlide} tasks. HER and HiS generates an order of magnitude more successful trajectories than HER or HiS on their own.}
\label{fig:succ_traj}
\end{figure}

A key reason why hindsight methods seem to work well is that they generate additional positively labeled experiences. 
To illustrate this phenomena, we look at a simple metric, the number of successful trajectories put into the RL buffer during the first 1000 episodes of training.
These trajectories are either collected on-policy or generated in hindsight if HiS and/or HER are involved.
We filter out trivial episodes that are labeled successful but where the object does not change position. Such episodes contain little useful information.
We note a strong correlation between the number of trajectories labeled successful in \fig{fig:succ_traj} and the results from \fig{fig:learning_curve_gym_rob}.
For both tasks, they show the same arrangement between the algorithms that we compare.
This finding illustrates well how HER and HiS complement each other.
For the two tasks studied, at the initial phases of the training, the combination of HER and HiS generates an order of magnitude more successful trajectories than HER or HiS individually.
The intuition behind this is that HiS generates a lot of hindsight trajectories, and HER labels them successful.

\subsection{Efficiency of HiS for Entirely Simulated Tasks}
\label{subsec:EntSim}

HiS was originally designed to speed up learning in a hybrid sim and real setting, where real data is expensive and simulated data is cheap. Therefore, the experiments so far, even those run only in simulation have been evaluated on the number of steps that would have been executed on a real robot.
But does it make sense to apply HiS in purely simulated tasks, e.g., in a sim-to-real setting for the simulation stage? In this setting, instead of minimizing the number of robot steps, the goal is usually to decrease total wall-clock time and computation. Compared to the baseline RL algorithm, HiS increases computation because it simulates extra virtual states, sorts this data according to the selection criterion and then feeds it into the replay buffer of the algorithm.
However, as shown earlier, HiS also increases sample efficiency during learning.

Tab.~\ref{tab:hissim} shows the evaluation of HiS regarding its efficiency for simulated tasks.
We compare HiS to vanilla SAC and the combination of HER and HiS to only HER.
To compare total computational cost, we restrict the learning to only one CPU (Intel Xeon W-2145 with 3.70GHz clock speed).
The effect of increased computation depends on the task and the number of additional virtual objects.
Wall-clock time for the same number of steps when using HiS increases only by about 2\% for the table tennis task but by about 60\% for the manipulation tasks.
To quantify improvements in sample efficiency, we look at the number of robot time steps HiS takes to reach the asymptotic or final performance of vanilla SAC and similarly the number of robot time steps HER and HiS takes to reach the asymptotic or final performance of HER.
As we have shown in \fig{fig:learning_curve_tt} and \fig{fig:learning_curve_gym_rob}, we achieve significant improvements by using HiS.
Therefore, combining these two effects, even when evaluated in wall-clock time instead of in robot time steps, HiS compares favorably. In conclusion, we have shown that HiS can be beneficial for settings where we want to optimize a policy only in simulation.

For this reason, HiS' range of applications extends beyond tasks that fulfill all the HySR assumptions as discussed in \sect{subsec:HySR}. 
In simulated environments, differences in the complexity of the dynamics of different parts of the task do not need to be considered.
However, HiS also assumes that the virtual state should not affect the real state. For purely simulated tasks, these effects onto the robot can be modeled, but a remaining issue is that multiple objects, that do not influence each other, can exert forces onto the robot simultaneously.
To overcome these discrepancies, one potential solution for tasks where contacts are sparse, could be to split the training into two stages: First, training with HiS and ignoring these discrepancies. This stage could help getting into contact with interesting objects. And later fine-tuning those contacts without HiS.

%% file: figures/plot_legend.tex
\begin{tikzpicture} 

  \definecolor{darkmagenta1604152}{RGB}{160,4,152}
  \definecolor{darkseagreen111194118}{RGB}{111,194,118}
  \definecolor{dodgerblue1101252}{RGB}{1,101,252}
  \definecolor{gainsboro}{RGB}{220,220,220}
  \definecolor{goldenrod21018910}{RGB}{210,189,10}
  \definecolor{lightgray}{RGB}{211,211,211}
  \definecolor{lightgray204}{RGB}{204,204,204}

    \begin{axis}[%
      legend columns=4,
    hide axis,
    width=3.4cm,
    xmin=0, xmax=1000,
    ymin=0.1, ymax=10000,
    legend style={
      fill opacity=1.0,
      draw opacity=1,
      text opacity=1,
      at={(0.5,1.18)},
      anchor=north,
      draw=lightgray204
    },
    ]

    \addlegendimage{very thick, darkseagreen111194118, opacity=1.0}
    \addlegendentry{HER + HiS}
    \addlegendimage{very thick, dodgerblue1101252, opacity=1.0}
    \addlegendentry{HiS}
    \addlegendimage{very thick, darkmagenta1604152, opacity=1.0}
    \addlegendentry{HER}
    \addlegendimage{very thick, goldenrod21018910, opacity=1.0}
    \addlegendentry{van. SAC}

    \end{axis}
\end{tikzpicture}

%% file: 5_conclusion.tex
In this work, we presented Hindsight States~(HiS), a method for sample-efficient training in a hybrid sim and real setup, where multiple instances of the virtual part of a task are paired with the single real part. 
HiS leverages this additional data by selectively choosing the data to train on.
We evaluated HiS on a variety of tasks in simulation and showed that it improves sample efficiency on a complex real-world muscular robot task.
We further showed that combining HER and HiS leads to even better performance than applying each method individually.

A limitation of our work is the requirement that the virtual part does not influence the real part. In many tasks, this influence cannot be neglected, particularly when dealing with heavier objects.
One way around, could be to map the forces from the simulation back onto the joint torques of a real robot.

Another strategy would be to split learning into multiple stages.
\begin{table}[h]
\centering
\caption{Evaluation of HiS regarding its efficiency for purely simulated tasks. There are two effects. First, using HiS alone or HiS in combination with HER increases total wall-clock time to perform the same number of time steps. However, secondly, on average HiS needs less of those time steps to reach matching performance (here: the asymptotic or final performance of SAC as evaluated during the experiments), and HER and HiS also need less time steps to reach matching performance (the asymptotic or final performance of HER as evaluated during the experiments).
For the tasks studied in this work, the second effect is larger, which results in less average wall-clock time to reach the same performance, even for learning purely in simulation.}
\label{tab:hissim}
\begin{tabular}{@{}rccc@{}}
\multicolumn{1}{c}{\multirow{2}{*}{}} & \multicolumn{3}{c}{Task} \\ \cmidrule(l){2-4} 
\multicolumn{1}{c}{} & Pushing & Sliding & Table Tennis \\ \midrule
\multicolumn{4}{c}{\textbf{Comparison of HiS to SAC}} \\ \midrule
total wall-clock time & 163.9~\% & 156.3~\% & 102.3~\% \\ \arrayrulecolor{black!30}\midrule
\begin{tabular}[c]{@{}r@{}}robot time steps for \\ matching performance\end{tabular} & 16.7~\% & 50.8~\% & 30.7~\% \\ \midrule
\begin{tabular}[c]{@{}r@{}}wall-clock time for \\ matching performance\end{tabular} & 27.4~\% & 79.4~\% & 31.4~\% \\ 
\arrayrulecolor{black}\midrule
\multicolumn{4}{c}{\textbf{Comparison of HER+HiS to HER}} \\ \midrule
total wall-clock time & 165.6~\% & 162.3~\% &  \\ \arrayrulecolor{black!30}\midrule
\begin{tabular}[c]{@{}r@{}}robot time steps for \\ matching performance\end{tabular} & 11.7~\% & 42.6~\% &  \\ \midrule
\begin{tabular}[c]{@{}r@{}}wall-clock time for \\ matching performance\end{tabular} & 19.4~\% & 69.1~\% &  \\ 
\end{tabular}
\end{table}
Stages where the condition is fulfilled could be learned with HiS and completely real training otherwise.
As an example, for a grasping task, the stage where the robot learns to get to the position just before the gripper touches the object could be learned using HiS.

For complex tasks that require long interactions, sampling virtual objects throughout the episode~(instead of only at the beginning) could produce better results. The advantage of this approach is that resampling objects close to the robot generates more important experiences with a higher chance due to additional contacts. 

The focus of this work was to devise a more sample efficient training scheme in the HySR setting.
However, as shown in \sect{subsec:EntSim}, HiS can also be beneficial for purely simulated tasks. Finding the best strategies to apply HiS to simulated tasks, especially to those that do not satisfy the HySR conditions holds a lot of potential.

Future work can also focus on combining HiS with curriculum learning by using data augmentations.
Data augmentations such as transformations as well as perturbations of the virtual objects, which itself can be physically plausible, facilitates generalization. 
In addition, one can adapt the laws of physics of the virtual part such as gravity or the speed of time to change the difficulty of the whole task. 
In this manner, a curriculum could be added that, e.g., slows down or accelerates time during the training, so the robot reaches high reward regions faster.
Subsequently, adjusting time gradually to its true speed toward the end of the training improves the transfer of the performance to the real world.

%% file: 99_appendix.tex
\subsection{Experiment Details}
\label{subsec:ExDetails}

In this section, we elaborate on the details of the tasks that we used for evaluating our method.

\subsubsection{Simulated Table Tennis Task}
We run our experiments in a custom MuJoCo environment, simulating the muscles with a Hill-type muscle model~\cite{haeufle_hill-type_2014}.
Similar to \citet{buchler_learning_2022}, we replay ball trajectories until contact and simulate afterwards. The average episode length is 37-39 time steps, depending on the policy. The episode ends when the ball touches the racket or gets out of reach of the racket.

For HiS, parallel episodes continue after the main episode ended. The problem is that we sample the policy distribution conditioned on the state containing the main virtual object.
In these cases, we sample the policy with one of the other virtual objects, until the episode is finished for all virtual objects. 

\subsubsection{Real Robot Table Tennis Task}

We keep most of the task details similar to the original experiment in~\citet{buchler_learning_2022} and list our adaptations. 
\begin{itemize}
    \item We employ our policy on a new iteration of the robot hardware.
    \item The initialization procedure is different: Instead of using a position controller, we send fixed target pressures to get the robot to an initial position.
    \item We send actions to the robot at a frequency of \SI{25}{\hertz} instead of \SI{100}{\hertz} because we found this to lead to significantly better results with SAC.
\end{itemize}
Collecting 15k episodes takes about 24 hours on the real robot.

\subsubsection{FetchPush and FetchSlide}

Each episode has a fixed length of 50 time steps. To apply HiS, we sample additional virtual objects according to the same distribution that is used to sample the main object.

\subsection{Hyperparameters}
\label{subsec:Hyperparameters}

The hyperparameters for the three tasks are shown in Tables~\ref{tab:hps_tt}, \ref{tab:hps_fetchpush}, and \ref{tab:hps_fetchslide}. The SAC hyperparameters for the table tennis task were tuned on the simulated task without HiS. The gym robotics experiments incorporate the hyperparameters found in~\cite{rl-zoo3}.

\renewcommand{\arraystretch}{1.2}       
\begin{table}
\caption{Hyperparameters table tennis}
	\label{tab:hps_tt}
	\begin{center}
    \begin{tabular}{c|l|lll}
                                                             & hyperparameter & value &  &  \\ \cline{1-3}
    \multicolumn{1}{c|}{\multirow{12}{*}{SAC}} & \multicolumn{1}{l|}{gamma}      & 0.9999     &  &  \\
    \multicolumn{1}{c|}{}                     & \multicolumn{1}{l|}{ent\_coef} & 0 &  &  \\
    \multicolumn{1}{c|}{}                     & \multicolumn{1}{l|}{learning\_rate} & 0.0003 &  &  \\
    \multicolumn{1}{c|}{}                     & \multicolumn{1}{l|}{batch\_size} & 256 &  &  \\
    \multicolumn{1}{c|}{}                     & \multicolumn{1}{l|}{policy\_network} & MLP &  &  \\
    \multicolumn{1}{c|}{}                     & \multicolumn{1}{l|}{num\_layers} & 1 &  &  \\
    \multicolumn{1}{c|}{}                     & \multicolumn{1}{l|}{num\_hidden} & 200 &  &  \\
    \multicolumn{1}{c|}{}                     & \multicolumn{1}{l|}{gradient\_steps} & 500 &  &  \\
    \multicolumn{1}{c|}{}                     & \multicolumn{1}{l|}{train\_freq} & 1 &  &  \\
    \multicolumn{1}{c|}{}                     & \multicolumn{1}{l|}{train\_freq\_unit} & episode &  &  \\
    \multicolumn{1}{c|}{}                     & \multicolumn{1}{l|}{buffer\_size} & 5000000 &  &  \\
    \multicolumn{1}{c|}{}                     & \multicolumn{1}{l|}{learning\_starts} & 10000 &  &  \\
    \cline{1-3}
    \multicolumn{1}{c|}{\multirow{3}{*}{HiS}} & \multicolumn{1}{l|}{criterion}  & reward per episode     &  &  \\
    \multicolumn{1}{c|}{}                     & \multicolumn{1}{l|}{$k_c$} & 3     &  &  \\
    \multicolumn{1}{c|}{}                     & \multicolumn{1}{l|}{$\psi_c$} & 0.5     &  &  \\
    \end{tabular}
    \end{center}
\end{table}

\begin{table}
\caption{Hyperparameters FetchPush}
	\label{tab:hps_fetchpush}
	\begin{center}
    \begin{tabular}{c|l|lll}
                                                             & hyperparameter & value &  &  \\ \cline{1-3}
    \multicolumn{1}{c|}{\multirow{12}{*}{SAC}} & \multicolumn{1}{l|}{gamma}      & 0.95     &  &  \\
    \multicolumn{1}{c|}{}                     & \multicolumn{1}{l|}{ent\_coef} & auto &  &  \\
    \multicolumn{1}{c|}{}                     & \multicolumn{1}{l|}{learning\_rate} & 0.001 &  &  \\
    \multicolumn{1}{c|}{}                     & \multicolumn{1}{l|}{batch\_size} & 256 &  &  \\
    \multicolumn{1}{c|}{}                     & \multicolumn{1}{l|}{policy\_network} & MLP &  &  \\
    \multicolumn{1}{c|}{}                     & \multicolumn{1}{l|}{num\_layers} & 2 &  &  \\
    \multicolumn{1}{c|}{}                     & \multicolumn{1}{l|}{num\_hidden} & 64 &  &  \\
    \multicolumn{1}{c|}{}                     & \multicolumn{1}{l|}{gradient\_steps} & 1 &  &  \\
    \multicolumn{1}{c|}{}                     & \multicolumn{1}{l|}{train\_freq} & 1 &  &  \\
    \multicolumn{1}{c|}{}                     & \multicolumn{1}{l|}{train\_freq\_unit} & step &  &  \\
    \multicolumn{1}{c|}{}                     & \multicolumn{1}{l|}{buffer\_size} & 5000000 &  &  \\
    \multicolumn{1}{c|}{}                     & \multicolumn{1}{l|}{learning\_starts} & 1000 &  &  \\
    \cline{1-3}
    \multicolumn{1}{c|}{\multirow{2}{*}{HER}} & \multicolumn{1}{l|}{goal\_selection\_strategy}  & future     &  &  \\
    \multicolumn{1}{c|}{}                     & \multicolumn{1}{l|}{n\_sampled\_goal} & 4     &  &  \\
    \cline{1-3}
    \multicolumn{1}{c|}{\multirow{3}{*}{HiS}} & \multicolumn{1}{l|}{criterion}  & $\Delta x$ virt. object     &  &  \\
    \multicolumn{1}{c|}{}                     & \multicolumn{1}{l|}{$k_c$} & 3     &  &  \\
    \multicolumn{1}{c|}{}                     & \multicolumn{1}{l|}{$\psi_c$} & 0.02     &  &  \\
    \end{tabular}
    \end{center}
\end{table}

\begin{table}
\caption{Hyperparameters FetchSlide}
	\label{tab:hps_fetchslide}
	\begin{center}
    \begin{tabular}{c|l|lll}
                                                             & hyperparameter & value &  &  \\ \cline{1-3}
    \multicolumn{1}{c|}{\multirow{12}{*}{SAC}} & \multicolumn{1}{l|}{gamma}      & 0.95     &  &  \\
    \multicolumn{1}{c|}{}                     & \multicolumn{1}{l|}{ent\_coef} & auto &  &  \\
    \multicolumn{1}{c|}{}                     & \multicolumn{1}{l|}{learning\_rate} & 0.001 &  &  \\
    \multicolumn{1}{c|}{}                     & \multicolumn{1}{l|}{batch\_size} & 2048 &  &  \\
    \multicolumn{1}{c|}{}                     & \multicolumn{1}{l|}{policy\_network} & MLP &  &  \\
    \multicolumn{1}{c|}{}                     & \multicolumn{1}{l|}{num\_layers} & 3 &  &  \\
    \multicolumn{1}{c|}{}                     & \multicolumn{1}{l|}{num\_hidden} & 512 &  &  \\
    \multicolumn{1}{c|}{}                     & \multicolumn{1}{l|}{gradient\_steps} & 1 &  &  \\
    \multicolumn{1}{c|}{}                     & \multicolumn{1}{l|}{train\_freq} & 1 &  &  \\
    \multicolumn{1}{c|}{}                     & \multicolumn{1}{l|}{train\_freq\_unit} & step &  &  \\
    \multicolumn{1}{c|}{}                     & \multicolumn{1}{l|}{buffer\_size} & 5000000 &  &  \\
    \multicolumn{1}{c|}{}                     & \multicolumn{1}{l|}{learning\_starts} & 1000 &  &  \\
    \cline{1-3}
    \multicolumn{1}{c|}{\multirow{2}{*}{HER}} & \multicolumn{1}{l|}{goal\_selection\_strategy}  & future     &  &  \\
    \multicolumn{1}{c|}{}                     & \multicolumn{1}{l|}{n\_sampled\_goal} & 4     &  &  \\
    \cline{1-3}
    \multicolumn{1}{c|}{\multirow{3}{*}{HiS}} & \multicolumn{1}{l|}{criterion}  & $\Delta x$ virt. object     &  &  \\
    \multicolumn{1}{c|}{}                     & \multicolumn{1}{l|}{$k_c$} & 3     &  &  \\
    \multicolumn{1}{c|}{}                     & \multicolumn{1}{l|}{$\psi_c$} & 0.02     &  &  \\
    \end{tabular}
    \end{center}
\end{table}

\subsection{Ablation Study: Selection Criterion}
\label{subsec:AblationSelect}

Using the reward criterion, which selects mainly successful trajectories, works well for HiS on all our tasks.
Because HER relabels trajectories as successful, it is less relevant to select successful HiS trajectories when applying HER and HiS together. Therefore, picking trajectories where the virtual object is moved, works even better in this case.
In this work, we focused on these criteria to showcase our method.
Our experiments indicate that the TD error criterion helps the learning progress, but less then the other two criteria. Table~\ref{tab:hiscriteria} shows an evaluation of the different selection criteria. Note that the real table tennis task was evaluated for only one training run. The simulated table tennis task was averaged over ten training runs with different random seeds, and the sliding task over five training runs. The table tennis task, both real and simulated, was evaluated after 15000 and the sliding task after 5000 episodes.

\begin{table}[h]
\centering
\caption{Evaluation of different selection criteria for HiS and HER+HiS on final success rate.}
\label{tab:hiscriteria}
\begin{tabular}{@{}rccc@{}}
\multicolumn{1}{c}{\multirow{2}{*}{}} & \multicolumn{3}{c}{Task} \\ \cmidrule(l){2-4} 
\multicolumn{1}{c}{} & Sim. Table Tennis & Real Table Tennis & Sliding\\ \midrule
van. SAC & 41.2~\% & 34.5~\% & 9.9~\% \\ \arrayrulecolor{black!30}\midrule
\begin{tabular}[c]{@{}r@{}}HiS reward\end{tabular} & 70.6~\% & 44.7~\% & 33.7~\% \\ \midrule
\begin{tabular}[c]{@{}r@{}}HiS $\Delta x$ virt. object\end{tabular} &  &  & 25.5~\% \\ \midrule
\begin{tabular}[c]{@{}r@{}}HiS TD error\end{tabular} & 50.8~\% & 34.2~\% & 13.2~\% \\ \midrule
\begin{tabular}[c]{@{}r@{}}HER+HIS reward\end{tabular} &  &  & 53.8~\% \\ \midrule
\begin{tabular}[c]{@{}r@{}}HER+HIS $\Delta x$ virt. object\end{tabular} &  &  & 64.5~\% \\ \midrule
\end{tabular}
\end{table}

%% file: main.bbl
\begin{thebibliography}{41}
\providecommand{\natexlab}[1]{#1}
\providecommand{\url}[1]{\texttt{#1}}
\expandafter\ifx\csname urlstyle\endcsname\relax
  \providecommand{\doi}[1]{doi: #1}\else
  \providecommand{\doi}{doi: \begingroup \urlstyle{rm}\Url}\fi

\bibitem[Andrychowicz et~al.(2017)Andrychowicz, Wolski, Ray, Schneider, Fong,
  Welinder, McGrew, Tobin, Pieter~Abbeel, and
  Zaremba]{andrychowicz_hindsight_2017}
Marcin Andrychowicz, Filip Wolski, Alex Ray, Jonas Schneider, Rachel Fong,
  Peter Welinder, Bob McGrew, Josh Tobin, OpenAI Pieter~Abbeel, and Wojciech
  Zaremba.
\newblock Hindsight experience replay.
\newblock \emph{Advances in Neural Information Processing Systems}, 30, 2017.

\bibitem[Bai et~al.(2019)Bai, Liu, Zhao, and Tang]{bai_guided_2019}
Chenjia Bai, Peng Liu, Wei Zhao, and Xianglong Tang.
\newblock Guided goal generation for hindsight multi-goal reinforcement
  learning.
\newblock \emph{Neurocomputing}, 359:\penalty0 353--367, 2019.

\bibitem[Beyene and Han(2022)]{beyene_prioritized_2022}
Sofanit~Wubeshet Beyene and Ji-Hyeong Han.
\newblock Prioritized hindsight with dual buffer for meta-reinforcement
  learning.
\newblock \emph{Electronics}, 11\penalty0 (24):\penalty0 4192, 2022.

\bibitem[Brittain et~al.(2019)Brittain, Bertram, Yang, and
  Wei]{brittain_prioritized_2019}
Marc Brittain, Josh Bertram, Xuxi Yang, and Peng Wei.
\newblock Prioritized sequence experience replay.
\newblock \emph{arXiv preprint arXiv:1905.12726}, 2019.

\bibitem[B{\"u}chler et~al.(2022)B{\"u}chler, Guist, Calandra, Berenz,
  Sch{\"o}lkopf, and Peters]{buchler_learning_2022}
Dieter B{\"u}chler, Simon Guist, Roberto Calandra, Vincent Berenz, Bernhard
  Sch{\"o}lkopf, and Jan Peters.
\newblock Learning to play table tennis from scratch using muscular robots.
\newblock \emph{IEEE Transactions on Robotics}, 38\penalty0 (6):\penalty0
  3850--3860, 2022.

\bibitem[B{\"u}chler et~al.(2023)B{\"u}chler, Calandra, and
  Peters]{buchler_learning_2023}
Dieter B{\"u}chler, Roberto Calandra, and Jan Peters.
\newblock Learning to control highly accelerated ballistic movements on
  muscular robots.
\newblock \emph{Robotics and Autonomous Systems}, 159:\penalty0 104230, 2023.

\bibitem[Büchler et~al.(2016)Büchler, Ott, and
  Peters]{buchler_lightweight_2016}
Dieter Büchler, Heiko Ott, and Jan Peters.
\newblock A {Lightweight} {Robotic} {Arm} with {Pneumatic} {Muscles} for
  {Robot} {Learning}.
\newblock In \emph{International {Conference} on {Robotics} and {Automation}
  ({ICRA})}, Stockholm, May 2016.
\newblock \doi{10.1109/icra.2016.7487599}.

\bibitem[Di-Castro et~al.(2021)Di-Castro, Di~Castro, and
  Mannor]{dicastro_sim_2021}
Shirli Di-Castro, Dotan Di~Castro, and Shie Mannor.
\newblock Sim and real: Better together.
\newblock \emph{Advances in Neural Information Processing Systems},
  34:\penalty0 6868--6880, 2021.

\bibitem[Fang et~al.(2018)Fang, Zhou, Shi, Gong, Xu, and Zhang]{fang_dher_2018}
Meng Fang, Cheng Zhou, Bei Shi, Boqing Gong, Jia Xu, and Tong Zhang.
\newblock {DHER}: Hindsight experience replay for dynamic goals.
\newblock In \emph{International Conference on Learning Representations},
  September 2018.

\bibitem[Fang et~al.(2019)Fang, Zhou, Du, Han, and Zhang]{fang_curriculum_2019}
Meng Fang, Tianyi Zhou, Yali Du, Lei Han, and Zhengyou Zhang.
\newblock Curriculum-guided hindsight experience replay.
\newblock \emph{Advances in Neural Information Processing Systems}, 32, 2019.

\bibitem[Fedus et~al.(2020)Fedus, Ramachandran, Agarwal, Bengio, Larochelle,
  Rowland, and Dabney]{fedus_revisiting_2020}
William Fedus, Prajit Ramachandran, Rishabh Agarwal, Yoshua Bengio, Hugo
  Larochelle, Mark Rowland, and Will Dabney.
\newblock Revisiting fundamentals of experience replay.
\newblock In \emph{International Conference on Machine Learning}, pages
  3061--3071. PMLR, 2020.

\bibitem[Haarnoja et~al.(2018)Haarnoja, Zhou, Abbeel, and
  Levine]{haarnoja_soft_2018}
Tuomas Haarnoja, Aurick Zhou, Pieter Abbeel, and Sergey Levine.
\newblock Soft {Actor}-{Critic}: Off-policy maximum entropy deep reinforcement
  learning with a stochastic actor.
\newblock In \emph{International Conference on Machine Learning}, pages
  1861--1870. PMLR, 2018.

\bibitem[Haeufle et~al.(2014)Haeufle, Günther, Bayer, and
  Schmitt]{haeufle_hill-type_2014}
D.~F.~B. Haeufle, M.~Günther, A.~Bayer, and S.~Schmitt.
\newblock Hill-type muscle model with serial damping and eccentric
  force–velocity relation.
\newblock \emph{Journal of Biomechanics}, 47\penalty0 (6):\penalty0 1531--1536,
  April 2014.
\newblock ISSN 0021-9290.
\newblock \doi{10.1016/j.jbiomech.2014.02.009}.

\bibitem[Higgins et~al.(2017)Higgins, Pal, Rusu, Matthey, Burgess, Pritzel,
  Botvinick, Blundell, and Lerchner]{higgins_darla_2017}
Irina Higgins, Arka Pal, Andrei Rusu, Loic Matthey, Christopher Burgess,
  Alexander Pritzel, Matthew Botvinick, Charles Blundell, and Alexander
  Lerchner.
\newblock Darla: Improving zero-shot transfer in reinforcement learning.
\newblock In \emph{International Conference on Machine Learning}, pages
  1480--1490. PMLR, 2017.

\bibitem[James et~al.(2019)James, Wohlhart, Kalakrishnan, Kalashnikov, Irpan,
  Ibarz, Levine, Hadsell, and Bousmalis]{james_sim_2019}
Stephen James, Paul Wohlhart, Mrinal Kalakrishnan, Dmitry Kalashnikov, Alex
  Irpan, Julian Ibarz, Sergey Levine, Raia Hadsell, and Konstantinos Bousmalis.
\newblock Sim-to-real via sim-to-sim: Data-efficient robotic grasping via
  randomized-to-canonical adaptation networks.
\newblock In \emph{Proceedings of the IEEE/CVF Conference on Computer Vision
  and Pattern Recognition}, pages 12627--12637, 2019.

\bibitem[Kang et~al.(2019)Kang, Belkhale, Kahn, Abbeel, and
  Levine]{kang_generalization_2019}
Katie Kang, Suneel Belkhale, Gregory Kahn, Pieter Abbeel, and Sergey Levine.
\newblock Generalization through simulation: Integrating simulated and real
  data into deep reinforcement learning for vision-based autonomous flight.
\newblock In \emph{International Conference on Robotics and Automation (ICRA)},
  pages 6008--6014. IEEE, 2019.

\bibitem[Kumar et~al.(2019)Kumar, Fu, Soh, Tucker, and
  Levine]{kumar_stabilizing_2019}
Aviral Kumar, Justin Fu, Matthew Soh, George Tucker, and Sergey Levine.
\newblock Stabilizing off-policy q-learning via bootstrapping error reduction.
\newblock \emph{Advances in Neural Information Processing Systems}, 32, 2019.

\bibitem[Kumar et~al.(2020)Kumar, Zhou, Tucker, and
  Levine]{kumar_conservative_2020}
Aviral Kumar, Aurick Zhou, George Tucker, and Sergey Levine.
\newblock Conservative {Q}-learning for offline reinforcement learning.
\newblock \emph{Advances in Neural Information Processing Systems},
  33:\penalty0 1179--1191, 2020.

\bibitem[Lange et~al.(2012)Lange, Gabel, and Riedmiller]{lange_batch_2012}
Sascha Lange, Thomas Gabel, and Martin Riedmiller.
\newblock Batch reinforcement learning.
\newblock \emph{Reinforcement learning: State-of-the-art}, pages 45--73, 2012.

\bibitem[Levine et~al.(2020)Levine, Kumar, Tucker, and Fu]{levine_offline_2020}
Sergey Levine, Aviral Kumar, George Tucker, and Justin Fu.
\newblock Offline reinforcement learning: Tutorial, review, and perspectives on
  open problems.
\newblock \emph{arXiv:2005.01643 [cs, stat]}, November 2020.
\newblock URL \url{http://arxiv.org/abs/2005.01643}.
\newblock arXiv: 2005.01643.

\bibitem[Lin(1992)]{lin_self-improving_1992}
Long-Ji Lin.
\newblock Self-improving reactive agents based on reinforcement learning,
  planning and teaching.
\newblock \emph{Machine learning}, 8\penalty0 (3):\penalty0 293--321, 1992.
\newblock Publisher: Springer.

\bibitem[Liu et~al.(2020)Liu, Bai, Zhao, Bai, Zhao, and
  Tang]{liu_generating_2020}
Peng Liu, Chenjia Bai, Yingnan Zhao, Chenyao Bai, Wei Zhao, and Xianglong Tang.
\newblock Generating attentive goals for prioritized hindsight reinforcement
  learning.
\newblock \emph{Knowledge-Based Systems}, 203:\penalty0 106140, 2020.

\bibitem[Mahler et~al.(2019)Mahler, Matl, Satish, Danielczuk, DeRose, McKinley,
  and Goldberg]{mahler_learning_2019}
Jeffrey Mahler, Matthew Matl, Vishal Satish, Michael Danielczuk, Bill DeRose,
  Stephen McKinley, and Ken Goldberg.
\newblock Learning ambidextrous robot grasping policies.
\newblock \emph{Science Robotics}, 4\penalty0 (26), 2019.

\bibitem[Mnih et~al.(2013)Mnih, Kavukcuoglu, Silver, Graves, Antonoglou,
  Wierstra, and Riedmiller]{mnih_playing_2013}
Volodymyr Mnih, Koray Kavukcuoglu, David Silver, Alex Graves, Ioannis
  Antonoglou, Daan Wierstra, and Martin Riedmiller.
\newblock Playing {Atari} with deep reinforcement learning.
\newblock \emph{arXiv preprint arXiv:1312.5602}, 2013.
\newblock URL \url{http://arxiv.org/abs/1312.5602}.

\bibitem[Moore and Atkeson(1993)]{moore_prioritized_1993}
Andrew~W. Moore and Christopher~G. Atkeson.
\newblock Prioritized sweeping: {Reinforcement} learning with less data and
  less time.
\newblock \emph{Mach Learn}, 13\penalty0 (1):\penalty0 103--130, October 1993.
\newblock ISSN 1573-0565.
\newblock \doi{10.1007/BF00993104}.
\newblock URL \url{https://doi.org/10.1007/BF00993104}.

\bibitem[Muratore et~al.(2022)Muratore, Ramos, Turk, Yu, Gienger, and
  Peters]{muratore_robot_2022}
Fabio Muratore, Fabio Ramos, Greg Turk, Wenhao Yu, Michael Gienger, and Jan
  Peters.
\newblock Robot learning from randomized simulations: A review.
\newblock \emph{Frontiers in Robotics and AI}, page~31, 2022.

\bibitem[Peng and Williams(1993)]{peng_efficient_1993}
Jing Peng and Ronald~J. Williams.
\newblock Efficient learning and planning within the {Dyna} framework.
\newblock \emph{Adaptive Behavior}, 1\penalty0 (4):\penalty0 437--454, March
  1993.
\newblock ISSN 1059-7123.
\newblock \doi{10.1177/105971239300100403}.
\newblock Publisher: SAGE Publications Ltd STM.

\bibitem[Peng et~al.(2018)Peng, Andrychowicz, Zaremba, and
  Abbeel]{peng_sim--real_2018}
X.~B. Peng, M.~Andrychowicz, W.~Zaremba, and P.~Abbeel.
\newblock Sim-to-real transfer of robotic control with dynamics randomization.
\newblock In \emph{2018 {IEEE} {International} {Conference} on {Robotics} and
  {Automation} ({ICRA})}, pages 1--8, May 2018.
\newblock \doi{10.1109/ICRA.2018.8460528}.

\bibitem[Rafailov et~al.(2021)Rafailov, Yu, Rajeswaran, and
  Finn]{rafailov_offline_2021}
Rafael Rafailov, Tianhe Yu, Aravind Rajeswaran, and Chelsea Finn.
\newblock Offline reinforcement learning from images with latent space models.
\newblock In \emph{Learning for Dynamics and Control}, pages 1154--1168. PMLR,
  2021.

\bibitem[Raffin(2020)]{rl-zoo3}
Antonin Raffin.
\newblock {RL} {Baselines3} {Zoo}.
\newblock \url{github.com/DLR-RM/rl-baselines3-zoo}, 2020.

\bibitem[Rauber et~al.(2019)Rauber, Ummadisingu, Mutz, and
  Schmidhuber]{rauber_hindsight_2019}
Paulo Rauber, Avinash Ummadisingu, Filipe Mutz, and J{\"u}rgen Schmidhuber.
\newblock Hindsight policy gradients.
\newblock \emph{arXiv:1711.06006 [cs]}, February 2019.
\newblock URL \url{http://arxiv.org/abs/1711.06006}.
\newblock arXiv: 1711.06006.

\bibitem[Ren et~al.(2019)Ren, Dong, Zhou, Liu, and Peng]{ren_exploration_2019}
Zhizhou Ren, Kefan Dong, Yuan Zhou, Qiang Liu, and Jian Peng.
\newblock Exploration via hindsight goal generation.
\newblock \emph{Advances in Neural Information Processing Systems}, 32, 2019.

\bibitem[Schaul et~al.(2015)Schaul, Quan, Antonoglou, and
  Silver]{schaul_prioritized_2015}
Tom Schaul, John Quan, Ioannis Antonoglou, and David Silver.
\newblock Prioritized experience replay.
\newblock \emph{arXiv preprint arXiv:1511.05952}, 2015.

\bibitem[Siegel et~al.(2020)Siegel, Springenberg, Berkenkamp, Abdolmaleki,
  Neunert, Lampe, Hafner, Heess, and Riedmiller]{siegel_keep_2020}
Noah~Y Siegel, Jost~Tobias Springenberg, Felix Berkenkamp, Abbas Abdolmaleki,
  Michael Neunert, Thomas Lampe, Roland Hafner, Nicolas Heess, and Martin
  Riedmiller.
\newblock Keep doing what worked: Behavioral modelling priors for offline
  reinforcement learning.
\newblock \emph{arXiv preprint arXiv:2002.08396}, 2020.

\bibitem[Todorov et~al.(2012)Todorov, Erez, and Tassa]{todorov_mujoco:_2012}
Emanuel Todorov, Tom Erez, and Yuval Tassa.
\newblock Mujoco: {A} physics engine for model-based control.
\newblock In \emph{Intelligent {Robots} and {Systems} ({IROS}), 2012
  {IEEE}/{RSJ} {International} {Conference} on}, pages 5026--5033. IEEE, 2012.

\bibitem[Wu et~al.(2019)Wu, Tucker, and Nachum]{wu_behavior_2019}
Yifan Wu, George Tucker, and Ofir Nachum.
\newblock Behavior regularized offline reinforcement learning.
\newblock \emph{arXiv preprint arXiv:1911.11361}, 2019.

\bibitem[Yu et~al.(2020)Yu, Thomas, Yu, Ermon, Zou, Levine, Finn, and
  Ma]{yu_mopo_2020}
Tianhe Yu, Garrett Thomas, Lantao Yu, Stefano Ermon, James Zou, Sergey Levine,
  Chelsea Finn, and Tengyu Ma.
\newblock {MOPO}: Model-based offline policy optimization.
\newblock \emph{arXiv:2005.13239 [cs, stat]}, May 2020.
\newblock URL \url{http://arxiv.org/abs/2005.13239}.
\newblock arXiv: 2005.13239.

\bibitem[Yu et~al.(2021)Yu, Kumar, Rafailov, Rajeswaran, Levine, and
  Finn]{yu_combo_2021}
Tianhe Yu, Aviral Kumar, Rafael Rafailov, Aravind Rajeswaran, Sergey Levine,
  and Chelsea Finn.
\newblock {COMBO}: Conservative offline model-based policy optimization.
\newblock \emph{Advances in Neural Information Processing Systems},
  34:\penalty0 28954--28967, 2021.

\bibitem[Zhang et~al.(2019)Zhang, Tai, Yun, Xiong, Liu, Boedecker, and
  Burgard]{zhang_vr_2019}
Jingwei Zhang, Lei Tai, Peng Yun, Yufeng Xiong, Ming Liu, Joschka Boedecker,
  and Wolfram Burgard.
\newblock {VR}-goggles for robots: Real-to-sim domain adaptation for visual
  control.
\newblock \emph{IEEE Robotics and Automation Letters}, 4\penalty0 (2):\penalty0
  1148--1155, 2019.

\bibitem[Zhao and Tresp(2018)]{zhao_energy_2018}
Rui Zhao and Volker Tresp.
\newblock Energy-based hindsight experience prioritization.
\newblock In \emph{Conference on Robot Learning}, pages 113--122. PMLR, 2018.

\bibitem[Zhu et~al.(2021)Zhu, Guo, Owaki, Kutsuzawa, and
  Hayashibe]{zhu_survey_2021}
Wei Zhu, Xian Guo, Dai Owaki, Kyo Kutsuzawa, and Mitsuhiro Hayashibe.
\newblock A survey of sim-to-real transfer techniques applied to reinforcement
  learning for bioinspired robots.
\newblock \emph{IEEE Transactions on Neural Networks and Learning Systems},
  pages 1--16, 2021.
\newblock ISSN 2162-2388.
\newblock \doi{10.1109/TNNLS.2021.3112718}.

\end{thebibliography}
